\documentclass[sn-mathphys,Numbered]{sn-jnl}

\usepackage{graphicx}%
\usepackage{multirow}%
\usepackage{amsmath,amssymb,amsfonts}%
\usepackage{amsthm}%
\usepackage{mathrsfs}%
\usepackage[title]{appendix}%
\usepackage{textcomp}%
\usepackage{manyfoot}%
\usepackage{booktabs}%
\usepackage{algorithm}%
\usepackage{algorithmicx}%
\usepackage{algpseudocode}%
\usepackage{listings}%
\usepackage{bm}
\usepackage{amsmath}
\usepackage{siunitx}
\usepackage{lineno}
\usepackage{hyperref}
\usepackage{float}
\usepackage[table]{xcolor}
\usepackage{hyperref}
\usepackage{float}
\usepackage[table]{xcolor}

\definecolor{level1}{RGB}{174,142,230}

\definecolor{level2}{RGB}{218,186,253}

\definecolor{level3}{RGB}{250,240,255}

\theoremstyle{thmstyleone}%
\theoremstyle{thmstyletwo}%

\theoremstyle{thmstylethree}%

\begin{document}

\title[Title]{Molecule Generation for Target Protein Binding with Hierarchical Consistency Diffusion Model}



\author[1]{\fnm{Guanlue} \sur{Li}}\email{guanlueli@gmail.com}
\equalcont{These authors contributed equally to this work.}

\author[4]{\fnm{Chenran} \sur{Jiang}}\email{jiangcr@szbl.ac.cn}
\equalcont{These authors contributed equally to this work.}

\author[1,3]{\fnm{Ziqi} \sur{Gao}}\email{zgaoat@connect.ust.hk}

\author[2]{\fnm{Yu} \sur{Liu}}\email{yliuil@connect.ust.hk}

\author[4]{\fnm{Chenyang} \sur{Liu}}\email{cyberyanglinox@outlook.com}

\author[4]{\fnm{Jiean} \sur{Chen}}\email{chenja@szbl.ac.cn}

\author*[2]{\fnm{Yong} \sur{Huang}}\email{yonghuang@ust.hk}

\author*[1,3]{\fnm{Jia} \sur{Li}}\email{jialee@ust.hk}

\affil[1]{\orgdiv{Data Science and Analytics}, \orgname{The Hong Kong University of Science and Technology}, \orgaddress{\city{Guangzhou}, \postcode{511400}, \country{China}}}

\affil[2]{\orgdiv{Department of Chemistry}, \orgname{The Hong Kong University of Science and Technology}, \orgaddress{\city{Hong Kong SAR}, \postcode{999077}, \country{China}}}

\affil[3]{\orgdiv{Division of Emerging Interdisciplinary Areas}, \orgname{The Hong Kong University of Science and Technology}, \orgaddress{\city{Hong Kong SAR}, \postcode{999077}, \country{China}}}

\affil[4]{\orgdiv{Pingshan Translational Medicine Center}, \orgname{Shenzhen Bay Laboratory}, \orgaddress{\city{Shenzhen}, \postcode{518118}, \country{China}}}



\abstract{Effective generation of molecular structures, or new chemical entities, that bind to target proteins is crucial for lead identification and optimization in drug discovery. Despite advancements in atom- and motif-wise deep learning models for 3D molecular generation, current methods often struggle with validity and reliability. To address these issues, we develop the Atom-Motif Consistency Diffusion Model (AMDiff), utilizing a joint-training paradigm for multi-view learning. This model features a hierarchical diffusion architecture that integrates both atom- and motif-level views of molecules, allowing for comprehensive exploration of complementary information. By leveraging classifier-free guidance and incorporating binding site features as conditional inputs, AMDiff ensures robust molecule generation across diverse targets. Compared to existing approaches, AMDiff exhibits superior validity and novelty in generating molecules tailored to fit various protein pockets. Case studies targeting protein kinases, including Anaplastic Lymphoma Kinase (ALK) and Cyclin-dependent kinase 4 (CDK4), demonstrate the model's capability in structure-based de novo drug design. Overall, AMDiff bridges the gap between atom-view and motif-view drug discovery and speeds up the process of target-aware molecular generation.}





\maketitle

\section{Introduction}\label{sec1}

In recent years, large-scale AI models have achieved remarkable breakthroughs, driving a surge of applications across various industries \cite{cumming2013chemical, hou2008structure}. The pharmaceutical sector, in particular, has benefited significantly, with tools like AlphaFold revolutionizing protein structure prediction \cite{jumper2021highly}. These advancements provide medicinal chemists with refined protein structures, accelerating structure-based drug design. Despite this progress, the chemical space of drug-like molecules remains vast and largely unexplored \cite{gomez2018automatic, polishchuk2013estimation}. Traditional methods, such as virtual screening, are often inefficient, costly, and limited to known structures. AI-based models offer promising capabilities to effectively navigate this chemical space. Consequently, developing novel AI tools for end-to-end, structure-based drug discovery has become a crucial research direction \cite{imrie2021deep, xu2021novo, isert2023structure}. In drug design, the interaction between a protein target and a ligand is often compared to the ``Lock and Key'' model, highlighting the necessity for precise structural complementarity \cite{tripathi2017molecular, xie2022advances}. The main challenge for medicinal chemists is to rapidly identify structurally novel and modifiable ``keys'' for the ``lock''—the target protein. Advanced AI methodologies have facilitated this process by accelerating lead generation. However, de novo molecule generation, guided by the target pocket, remains insufficiently developed \cite{anstine2023generative, fromer2023computer}. Unlike the straightforward ``Lock and Key'' analogy, this task requires processing 3D information within a flexible, continuous space. The limited data available often results in discontinuities and inaccuracies in atomic arrangements, leading to deviations from real-world atomic connectivity rules and bond lengths and angles that do not align with energy principles.

Existing models for de novo molecular generation often draw inspiration from real-life lead optimization strategies used in drug discovery, primarily chemical derivatization \cite{garcia2007chemical} and scaffold hopping \cite{zhao2007scaffold}. Chemical derivatization involves a sequential approach, where molecules branch out from a known starting point. In contrast, scaffold hopping retains the molecule's overall 3D shape while altering atom connectivity. Recent tools like GraphBP \cite{liu2022generating} and FLAG \cite{zhang2022molecule} implement chemical derivatization by sequentially introducing specific atoms or motifs into a binding site. Meanwhile, ScaffoldGVAE \cite{hu2023scaffoldgvae} employs scaffold hopping by preserving side chains and modifying the main core. Additionally, novel frameworks, such as one-shot generation methods, present intriguing possibilities by creating entire molecular structures simultaneously \cite{zhu2022survey}. TargetDiff \cite{guan20233d} and DecompDiff \cite{guan2024decompdiff} utilize this approach, employing diffusion models to generate molecules at the atom level in a single step. Regardless various strategies, these models typically use either individual atoms or motifs as building blocks for molecule construction. Atom-based de novo drug design methods \cite{zhang2023learning, zhang2023resgen, lin2022diffbp} are not limited by predefined motif libraries, allowing exploration of vast chemical spaces and generation of highly diverse compounds. Yet, these methods are confronted with 

\begin{figure}[H]%
\centering
\includegraphics[width=1\textwidth]{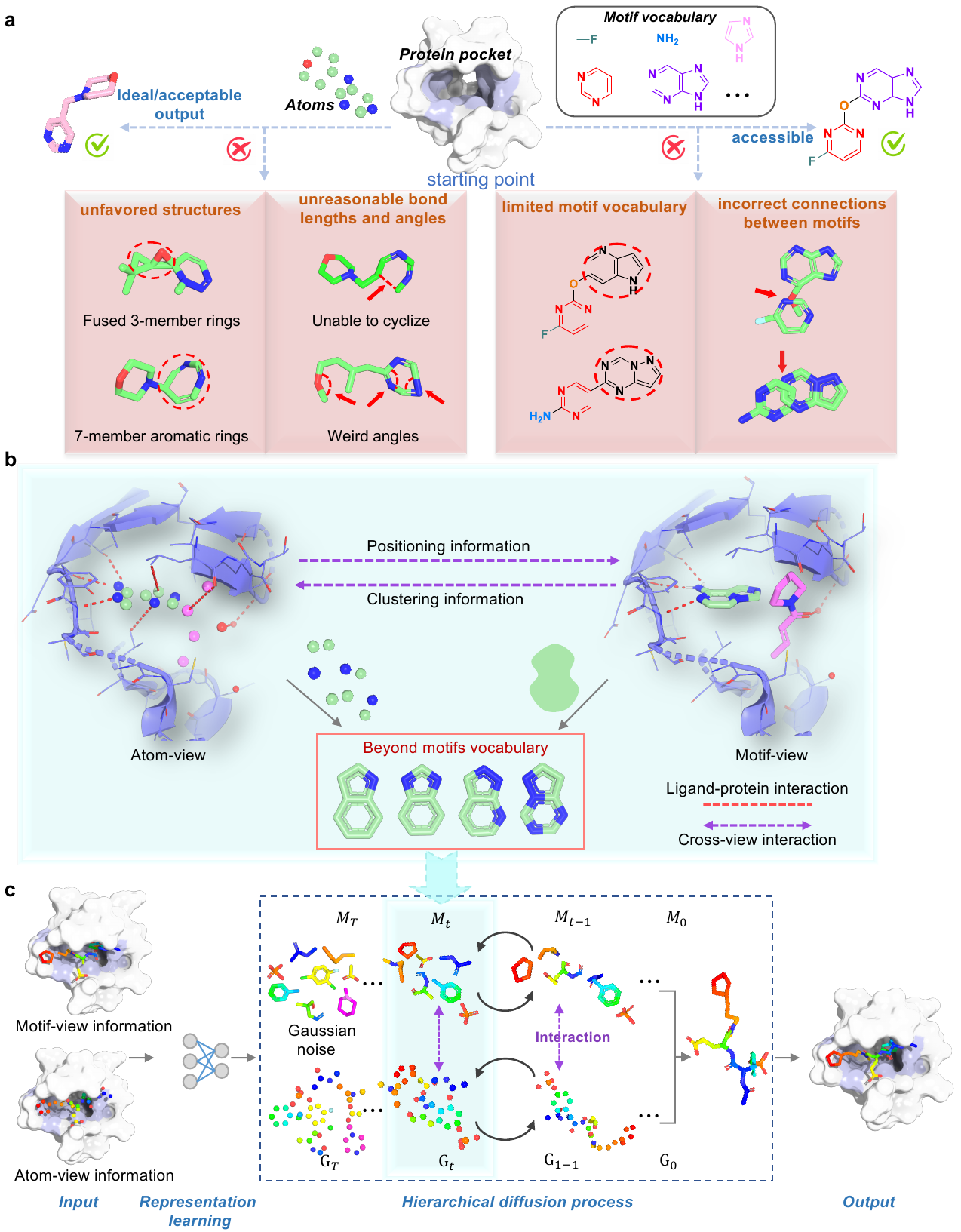}
\caption{(a) Ideal outputs and disadvantages of atom-based and motif-based methods for structure-based drug design. In atom-based methods (on the left), individual atoms serve as the fundamental units to construct highly diverse molecular structures. While these methods excel in generating variety, they often struggle to maintain coherence and realism in substructure formation, frequently leading to the creation of bonds with incorrect lengths and angles. Moreover, atom-based approaches can inadvertently produce unstable configurations such as three-membered rings. Conversely, motif-based methods (on the right) utilize predefined building blocks sourced from a motif vocabulary derived from existing datasets and chemical knowledge. However, these methods face limitations when desired motifs, such as 1H-pyrrolo[3,2-b]pyridine and pyrazolo[1,5-a][1,3,5]triazine, are absent from the vocabulary, potentially limiting structural diversity. Additionally, conflicts may arise in connecting different motifs, posing further challenges in generating cohesive structures. (b) The illustration of hierarchical-interaction information for ligand generation in this work. A ligand is decomposed into atoms and motifs, respectively. In the atom-view and motif-view, interaction details between the ligand and protein (represented by red dotted lines) are gathered using dedicated message passing networks. Additionally, for cross-view interactions (indicated by purple dashed line), facilitate the exchange of clustering and positioning information between the atom and motif views. (c) The AMDiff architecture is a diffusion-based model for hierarchical molecular generation. The AMDiff architecture is centered on a diffusion model that integrates atom-view and motif-view perspectives, each crucial for molecular generation. This model employs a conditional diffusion approach to recover noisy molecular structures and generate new ones through interactive denoising. In the atom-view, the model predicts atom types and positions, while in the motif-view, it constructs motif trees and generates predictions based on them. This architectural design fosters effective information exchange between views, providing valuable insights across various granularity levels in molecular structures. 
}\label{overview1}
\end{figure}

\noindent the validity of bond lengths and angles, which can result in the formation of structurally bizarre molecules. In contrast, motif-based approaches  \cite{powers2023geometric, zhang2022molecule, wills2023fragment} utilize predefined libraries to assemble molecules, but the reliance on existing datasets and current chemical knowledge limits exploration of unknown chemical spaces. This restriction confines the potential to generate novel structures beyond the available fragments.

To balance novelty and validity, a hierarchical graph model can be used to generate molecules simultaneously at both the atom and motif levels. Several pioneering works have been inspired by this multi-granularity modeling. DrugGPS \cite{zhang2023learning} incorporates an intrinsic two-level structure of the protein, featuring atom-level and residue-level encoders to learn sub-pocket prototypes, generating molecules motif-by-motif. Our development, HIGH-PPI \cite{gao2023hierarchical} aims to establish a robust understanding of Protein-Protein Interactions (PPIs) by creating a hierarchical graph that includes both the PPI graph and the protein graph. 
To achieve hierarchical performance at both the atom and motif levels, we introduce the Atom-Motif Consistency Diffusion Model (AMDiff), designed to efficiently generate high-quality 3D molecules for specific binding targets. AMDiff learns target information and constructs a graph structure incorporating topological details. At the ligand level, it employs a hierarchical diffusion approach, capturing both atom-view and motif-view of molecules to fully utilize available information. During molecular generation, we ensure that samples from both views are closely aligned in chemical space. The motif view provides insights into prior patterns, like aromatic rings, which the atom view might miss, while the atom view models diverse structures without being constrained by predefined motifs. The joint training approach leverages complementary information from different views, enhancing interaction during training. 
AMDiff employs the classifier-free guidance diffusion model in each view. We incorporate features extracted from binding sites as conditional inputs and train both conditional and unconditional diffusion models by randomly omitting the conditioning. This approach ensures balanced molecule synthesis across multiple targets.
To enhance the coherence and connectivity of generated molecules, we incorporate persistent homology, a technique from topological data analysis (TDA). This method captures multiscale topological features from both molecules and binding sites. By integrating these topological features, we strengthen the structural characteristics of the generated molecules and refine binding site topology identification based on shape properties.
We apply AMDiff to banchmark dataset and two kinase targets, demonstrating superior generation performance and effectiveness compared to other models. AMDiff exhibits exceptional performance when benchmarked to baseline methods, encompassing both atom- and motif-alone models, across diverse metrics. Further analysis on its robustness has verified that AMDiff can produce compounds tailored to varying pocket sizes.

\vspace{-3mm}
\section{Results}\label{sec3}

\subsection{Atom-based and motif-based methods are complementary in target-aware molecule generation}

Considering that molecular structures can be broken down into multiple levels of resolution, we aim to fully harness the potential of multi-level molecular structure generation. To this end, we propose AMDiff, a classifier-free hierarchical diffusion model designed for accurate and interpretable de novo ligand design through atom-motif consistency. 
Traditional ligand design strategies often utilize either individual atoms or motifs as foundational elements. Each granularity level offers distinct advantages and mechanisms for establishing interactions within the pocket environment, and they are not interchangeable. Depending solely on one resolution level may inadvertently overlook crucial structural patterns present at the other level. On one hand, using atoms as foundational units offers flexibility in accessing all possible coordinates in 3D space. However, it is challenging to produce reasonable structures due to the lack of necessary constraints to obey the fundamental electronic, steric, and geometric principles of chemistry. Atom-based methods often result in structural errors, such as unrealistic bond lengths and angles, which hinder proper ring formation. These models frequently generate thermodynamically unstable structures, like those with unstable fused cyclopropanes and seven-membered aromatic rings, due to the absence of Euclidean geometric constraints (Figure \ref{overview1}, left panel). On the other hand, motif-based design builds a motif vocabulary from existing drug datasets and chemical knowledge, selecting suitable motifs to assemble final molecules. However, this approach faces limitations due to the restricted motifs contained in the vocabulary, which hinders access to structures that incorporate fragments beyond the existing motif repertoire, as demonstrated in Figure \ref{overview1} (a), the right-hand panel. Additionally, incorrect integration between motifs can occur, such as improper linker construction or missing connections. These challenges are similar to those faced in atom-based generation approaches (For detailed comparison, refer to Table \ref{tab2}). 

In the proposed model, AMDiff operates hierarchically, incorporating both atomic and motif views, as illustrated in Figure \ref{overview1} (b). To connect these two views effectively, an interaction network is utilized. This network facilitates the exchange of complementary information between the atom-view and motif-view, enhancing the overall model performance. We establish ligand-protein interactions and cross-view interactions. Ligand-protein interactions are modeled through an equivariant graph neural network, which ensures that the generated molecules fit the target binding sites accurately by considering both geometric and chemical properties. Moreover, cross-view interactions are constructed to bridge the gap between atom-level precision and motif-level abstraction. Motifs interact with the target pocket, offering clustering information to the atom view, while the atom view provides detailed positioning information to the motif view. This bidirectional flow of information ensures that the generated ligands not only fit the binding sites but also maintain structural coherence beyond the predefined motif vocabulary. A schematic view of the AMDiff architecture is shown in Figure \ref{overview1} (c). The initial step in AMDiff involves obtaining representations of the protein and ligand through an embedding network. Subsequently, a denoising network predicts the state of ligand without noise. 
Each view employs a denoising process to predict the structure conditioned on binding sites, which includes a forward chain that perturbs data to noise and a reverse chain that converts noise back to data.  In the atom-view, The model focuses on is on capturing the fine-grained details of atomic positions and interactions. This involves learning the precise atomic-level forces and positional information, providing a broader context that aids in forming reasonable molecular clusters and overall topology. In the motif-view, the model captures higher-level structural patterns, such as functional groups and larger molecular fragments, ensuring that the generated ligands are structurally coherent and chemically valid. By obtaining the persistence diagram and encoding it as topological fingerprints, AMDiff effectively captures the multi-scale topological features essential for accurate ligand generation, as detailed in Section \ref{model_Hierarchical}.

We train our model on the CrossDocked dataset \cite{francoeur2020three}. During the training phase, both atom-view and motif-view particles, along with their corresponding binding protein pockets, are input to the model. The protein pocket remains fixed as it serves as the conditional information. In the sampling stage, we initialize the data state by sampling from a standard normal distribution, $\mathcal{N}(0,\mathbf{I})$. Subsequent states are iteratively generated using $p_{\theta}(G_{t-1}|G_{t}, C)$, where $C$ represents the condition. We evaluate AMDiff on the CrossDocked dataset, as well as on Anaplastic Lymphoma Kinase (ALK) and Cyclin-dependent kinase 4 (CDK 4) targets.
We assess the performance of our model from two perspectives: (1) Understanding the characteristic property distributions of ligands in different protein pockets. This entails learning the interaction patterns with protein pockets in order to achieve stronger binding. (2) Generating molecules for real-world therapeutic targets and exploring their interactions in the presence of mutated target proteins and varying pocket sizes. 

\subsection{AMDiff shows the best performance and generalization}\label{subsec32}

We conduct a comprehensive evaluation of AMDiff's performance in generating molecular structures. Specifically, we select 100 protein targets and generate 100 molecules for each target, resulting in a total of 10,000 generated molecules. We compare the result with recent atom-based methods including LiGAN, AR, Pocket2Mol, GraphBP, DecompDiff and the motif-based FLAG. The evaluation metrics for molecule generation performance include Diversity, Novelty, QED, SA, and Affinity, as defined in Section \ref{model_evalutation}.

\begin{table*}[tb]
\caption{Comparison of molecular properties between the baseline models and AMDiff targeting the CrossDocked \cite{francoeur2020three} test set. The top three performing models are denoted by distinct colors, with the highest-performing model indicated by the darkest purple background color.}\label{tab1}%
\resizebox{\textwidth}{15mm}{
\begin{tabular}{@{}llllllll@{}}
\toprule
\textbf{Model}  & \textbf{Validity $\uparrow$}   & \textbf{Diversity $\uparrow$}  & \textbf{Novelty $\uparrow$} & 
\textbf{QED $\uparrow$} & \textbf{SA $\uparrow$} & \textbf{Affinity $\downarrow$}   \\
\midrule

\textbf{liGAN}   &   0.950 $\pm$ 0.021    &   0.488 $\pm$ 0.043    &   0.304 $\pm$ 0.016    &    0.411 $\pm$ 0.103   &   0.590 $\pm$ 0.089  &    -6.331 $\pm$ 1.108   &  \\
\textbf{AR}             &   0.843 $\pm$ 0.012    &    0.583 $\pm$ 0.145   &   0.490 $\pm$ 0.177    &        0.434 $\pm$ 0.018 &  \cellcolor{level2} 0.676 $\pm$ 0.170   &    -6.231 $\pm$ 1.660    &  \\
\textbf{Pocket2Mol}     &   0.958 $\pm$ 0.002    &  \cellcolor{level3} 0.612 $\pm$ 0.035    &   \cellcolor{level3} 0.558 $\pm$ 0.102    &    0.454 $\pm$ 0.165   &   \cellcolor{level3} 0.629 $\pm$ 0.134  &    -7.269 $\pm$  2.239   &  \\
\textbf{GraphBP} &  \cellcolor{level3}   0.974 $\pm$ 0.004    &   0.594 $\pm$ 0.014    &   0.547 $\pm$ 0.007    &    0.439 $\pm$ 0.090   &   0.512 $\pm$ 0.134  &   \cellcolor{level2} -7.381 $\pm$ 2.101    &  \\
\textbf{DecompDiff}  &  0.967  $\pm$ 0.016  &   \cellcolor{level2} 0.668  $\pm$  0.026  &  \cellcolor{level2} 0.624  $\pm$ 0.073   & \cellcolor{level3}  0.461  $\pm$ 0.154  &  0.624  $\pm$  0.131 &  \cellcolor{level3} -7.324  $\pm$  0.211   &  \\
\textbf{FlAG}  &   \cellcolor{level2}  0.981 $\pm$ 0.013   &    0.604 $\pm$ 0.024   &    0.499 $\pm$ 0.129   &   \cellcolor{level2} 0.463 ± 0.129    &   0.508 $\pm$ 0.150   &   -7.169 $\pm$ 2.019     &  \\

\textbf{AMDiff (ours)}   & \cellcolor{level1} 0.989 $\pm$  0.007   &   \cellcolor{level1} 0.672 $\pm$ 0.013   &   \cellcolor{level1}  0.663 $\pm$ 0.104   &   \cellcolor{level1}  0.479  $\pm$ 0.209   &   \cellcolor{level1} 0.684 $\pm$ 0.125  &  \cellcolor{level1}   -7.466 $\pm$ 2.062    &  \\
\textbf{Testset}   &     -    &   -    &   -    &   0.476 $\pm$ 0.206   &  0.727 $\pm$ 0.140  &  -7.502 $\pm$ 1.898    &  \\
\botrule
\end{tabular}}
\end{table*}

In table \ref{tab1}, we show the mean values with standard deviations of evaluation metrics. Generally, our method demonstrates best performance compared to the baseline methods. AMDiff achieves a 98.9\% output validity, indicating its ability to accurately learn the chemical constraints of topological structures. The higher percentages of diversity and novelty in the generated molecules indicate that our model effectively explores the chemical space beyond the molecular structures present in the training dataset. We assess the affinity of the generated conformations of molecules by calculating the Vina docking value between the molecules and target proteins. Compared to the second-best method, AMDiff improves the QED ratio by 5.5\%. Among the baseline methods, our model achieves the highest SA score. 
The results in Table \ref{tab1} indicate that our model achieves an average affinity of -7.466 kcal/mol, demonstrating the model's capability to generate molecules with favorable binding affinity. Overall, our model exhibits improved performance compared to other methods.

\begin{figure*}[tb]%
\centering
\includegraphics[width=1\textwidth]{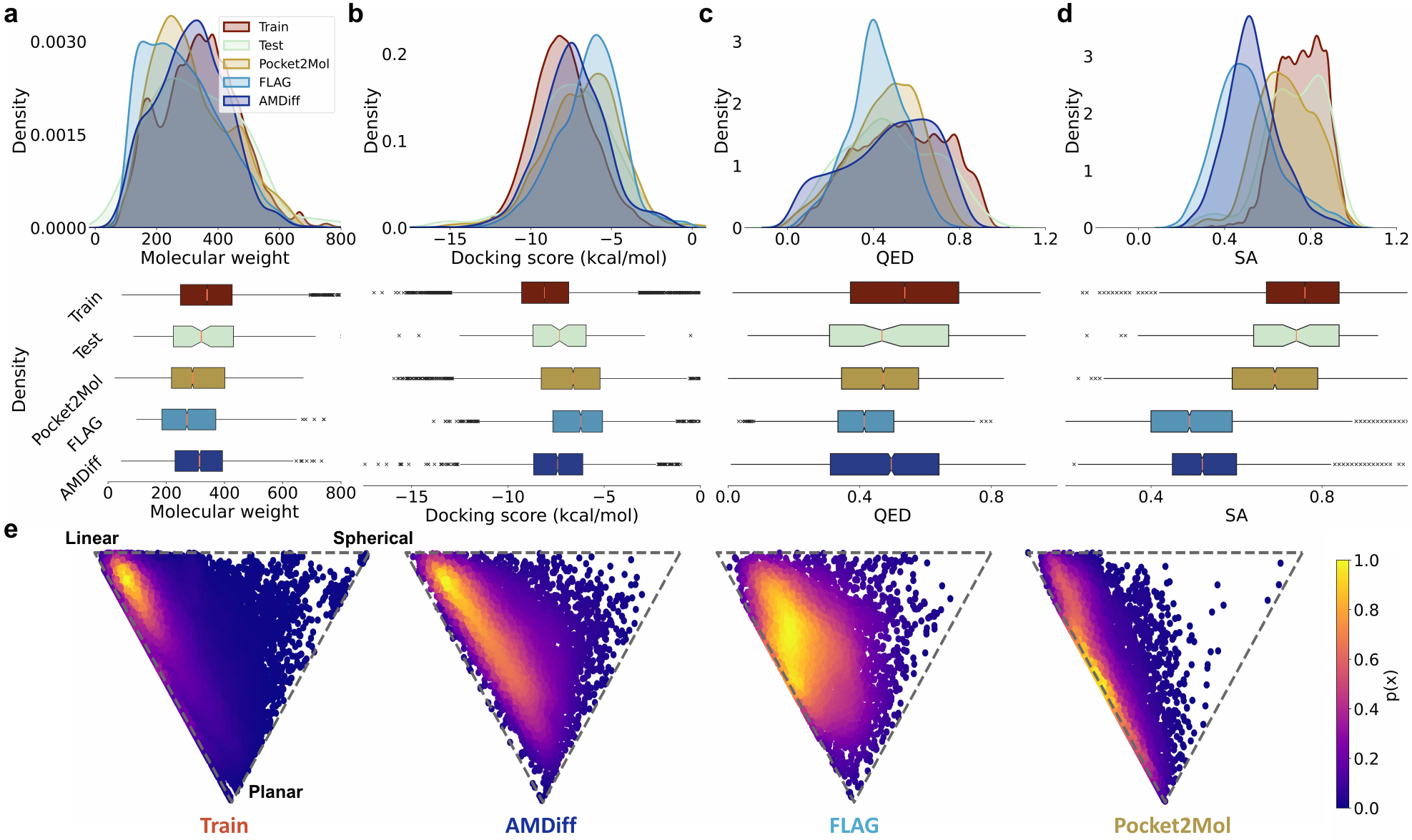}
\caption{Quantitative evaluations of the models targeting the CrossDocked \cite{francoeur2020three} test set. (a-d) The distribution of the following metrics: \textbf{(a)} Docking score; \textbf{(b)} Molecular weight; \textbf{(c)} QED; \textbf{(d)} SA, comparing AMDiff (purple), Pocket2Mol (blue), FLAG (green), Train set (red), and Test set (yellow) molecules. \textbf{(e)} visualizes the 3D shape distribution of the generated molecules using NPR descriptors.}\label{result1}
\end{figure*}

To further evaluate AMDiff's capacity to accurately capture the distribution of training dataset and detect the distribution shifts with the test dataset, we conduct an additional analysis focusing on the physicochemical properties and topological structures of the generated molecules. The distribution patterns of various key metrics, including docking score, molecular weight, QED, and SA, are presented in Figure \ref{result1}. The result shows that the generated molecules exhibit lower docking scores and smaller standard deviations compared to other baselines, indicating a higher affinity with target proteins. Regarding molecular weight, our model closely resembles that of known ligands, outperforming Pocket2Mol and FLAG models. When comparing the QED and SA distribution respectively, our model demonstrates mean values closer to the training and test sets. To better understand the chemical space occupied by the generated molecules, we assess the 3D shape distribution using NPR descriptors.
As shown in Figure \ref{result1} (e), the molecules generated by our model tend towards a linear shape, slightly leaning towards the ``planar" corner. This alignment with the training set suggests that our model produces molecules consistent with reference ligands. Conversely, both Pocket2Mol and FLAG models exhibit more divergent distributions, indicating a deviation from the expected shape characteristics, whose center of the distribution is positioned apart from the center of bioactive ligands. Through above quantitative evaluations, AMDiff demonstrates excellent results by generating diverse and active results molecules, outperforming other methods in the drug design process.

We also calculate various bond angles and dihedral angle distributions for the generated molecules and compare them against the respective reference empirical distributions using Kullback-Leibler (KL) divergence. Bond angles and dihedral angles describe the spatial arrangement between three connected atoms and the rotation between planes defined by four sequential atoms, respectively.
As depicted in Table \ref{tab2}, our model demonstrates lower KL divergence compared to all other atom-based baselines. Our model show competetive performance with FLAG, which operates on a motif-wise basis. The motif-wise models offer the advantage of predicting more precise angles within motifs by employing a strategy of combining predefined motifs from an established vocabulary. However, motif-based models also face the challenge of constructing cohesive connections between motifs.
The results highlight the effectiveness of AMDiff in capturing geometric characteristics and realistic substructures of bond angles and dihedral angles by utilizing atom and motif-views, thereby approaching the performance of motif-based models without relying on a predefined substructures.   

\begin{table*}[tb]
\caption{The KL divergence measures the difference in bond angles and dihedral angles between the reference molecules and the generated molecules targeting the CrossDocked \cite{francoeur2020three} test set. The lowercase letters represent the atoms in the aromatic rings. Lower values indicate that the models better capture the distribution of realistic structures. All the baseline models are atom-wise, with the exception of FLAG, which is a motif-based model. The top three performing models are denoted by distinct colors, with the highest-performing model indicated by the darkest purple background color.}\label{tab2}%
\centering
\resizebox{\textwidth}{16mm}{
\begin{tabular}{@{}l|lllll|lllll@{}}
\toprule
\textbf{Model}  & \textbf{CCC} &  \textbf{CCO} & \textbf{CNC}  & \textbf{CCN} & \textbf{CC=O} & \textbf{CCCC} & \textbf{CNCC} & \textbf{NCCN} & \textbf{CC=CC} & \textbf{CCCS}  \\
\midrule
\textbf{liGAN} & 6.845 & 8.314  &  6.737 & 5.672 &   6.752   &  1.479  &   2.355 &   3.689 &  5.216  & 1.229   \\
\textbf{AR} & 1.973 & 1.857 & 2.54 & 2.361 &  2.743  &  1.271  &  0.947 & 1.417 & 3.575 &  0.854    \\
\textbf{Pocket2Mol} & 0.917 & 0.874   & 0.465  &  0.562  &   0.765  & 0.954  & 0.864 & \cellcolor{level2} 0.368 & 2.177 & 0.386      \\
\textbf{GraphBP} & 0.573 & 0.472   &  0.341  & \cellcolor{level3} 0.304  & 0.458  & \cellcolor{level3} 0.896  & 1.027 &0.876 & \cellcolor{level3} 0.648 & 0.351       \\
\textbf{DecompDiff}    &\cellcolor{level3} 0.321 & \cellcolor{level3} 0.302 & \cellcolor{level3} 0.286  & 0.325     & \cellcolor{level3} 0.244 & 1.056 &  \cellcolor{level2} 0.566  & \cellcolor{level3} 0.437 & 1.573 & \cellcolor{level2} 0.296    \\
\textbf{FLAG}   & \cellcolor{level1} 0.251     & \cellcolor{level1}  0.274      & \cellcolor{level1} 0.196 &\cellcolor{level1} 0.287   &   \cellcolor{level2} 0.216   & \cellcolor{level1} 0.642  & \cellcolor{level3} 0.658 & 0.547 & \cellcolor{level2} 0.336 &  \cellcolor{level3} 0.317  \\
\textbf{AMDiff} & \cellcolor{level2}  0.306   & \cellcolor{level2} 0.297    &\cellcolor{level2} 0.257  &\cellcolor{level2} 0.294   & 
\cellcolor{level1} 0.205 & \cellcolor{level2} 0.863    &  \cellcolor{level1} 0.469 &\cellcolor{level1} 0.348 & \cellcolor{level1} 0.292 & \cellcolor{level1} 0.208  \\

\botrule
\end{tabular}}
\end{table*}
 
\subsection{AMDiff enables target-aware molecule generation for kinase targets}\label{subsec33}

Our study focuses on harnessing the potential of AMDiff to predict novel small molecule binders to drug targets, aiming to expedite lead identification and optimization processes. Beyond examining the statistical performance metrics, our evaluation delves deep into the practical effectiveness of the model-generated molecules against real drug targets. In this aspect, we conduct a thorough assessment targeting two significant therapeutic kinases: ALK and CDK4. ALK, a receptor tyrosine kinase, plays a pivotal role in driving the progression of specific cancers like non-small cell lung cancer (NSCLC) and anaplastic large cell lymphoma (ALCL) when mutated or rearranged. Inhibitors of ALK disrupt the aberrant signaling pathways activated by mutated ALK proteins, effectively impeding cancer cell proliferation and viability. Notably, several ALK inhibitors, including crizotinib, ceritinib, alectinib, and lorlatinib, have been approved for treating NSCLC. On the other hand, CDK4, a key regulator of the cell cycle involved in the G1 to S phase transition, exerts control over cell proliferation by phosphorylating the retinoblastoma protein (Rb), leading to the release of E2F transcription factors. By inhibiting CDK4, drugs can halt the hyperphosphorylation of Rb, preserving its growth-suppressive function and arresting the cell cycle at the G1 phase. This mechanism underscores the potential of CDK4 inhibitors as treatments for cancers characterized by disrupted cell cycle regulation, such as breast cancer, melanoma, and certain sarcomas.

\begin{figure*}[tb]%
\centering
\includegraphics[width=1\textwidth]{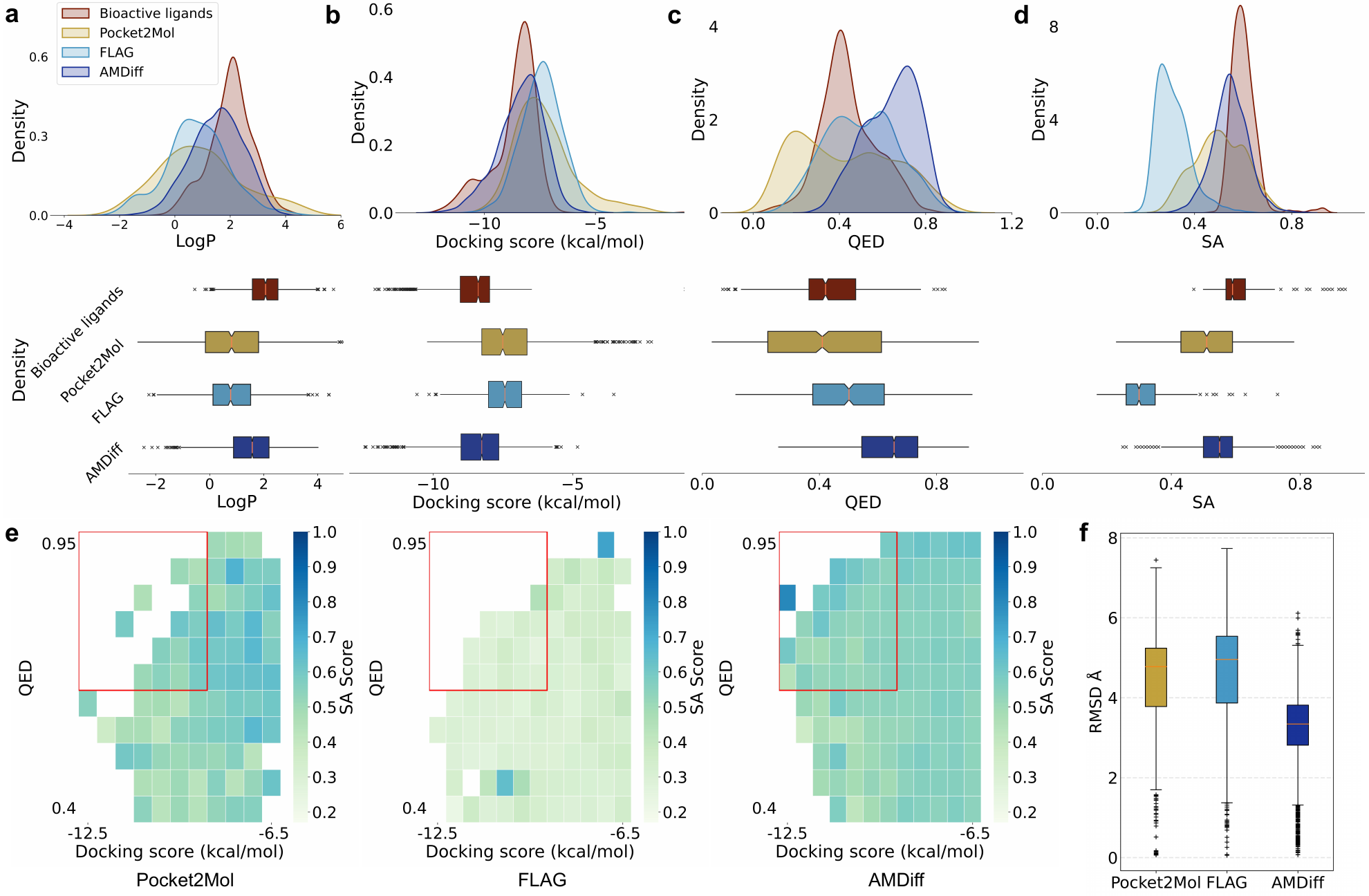}
\caption{Quantitative evaluations of the models targeting ALK (PDB id: 3LCS). The distributions of the following metrics were analyzed: \textbf{(a)} Docking score; \textbf{(b)} Molecular weight; \textbf{(c)} QED; \textbf{(d)} LogP, comparing the performance of AMDiff (purple), Pocket2Mol (yellow), FLAG (green) models, molecules, and bioactive ligands (red). \textbf{(e)} The distribution of Docking score, QED, and SA score for the generated samples was visualized. The drug-like region with QED $\geq$ 0.65 and docking score $\leq$ -8.5 (kcal/mol) is indicated with red boxes. \textbf{(f)} the RMSD was calculated to determine the conformational changes before and after the docking process.}\label{result2}
\end{figure*}

Following previous medicinal chemistry efforts \cite{pan2017combating,jiang2024elucidating}, we focus on the ATP-binding pockets of these two proteins, executed large-scale molecular generation, and systematically summarized and compared the performance of the generated molecules across various parameters. We employ AMDiff to generate 15,000 molecules and utilized a molecular filter, as described in Section \ref{model_MCF}, to identify high-quality candidates. 
Figure \ref{result2} (a) to (d) illustrates the affinity and drug-likeness properties of molecules targeting ALK (PDB id: 3LCS) produced by different methods. A bioactive ligand dataset serves as the reference, establishing a baseline for models capable of designing ALK-targeted active-like molecules.
\begin{figure*}[tb]%
\centering
\includegraphics[width=1\textwidth]{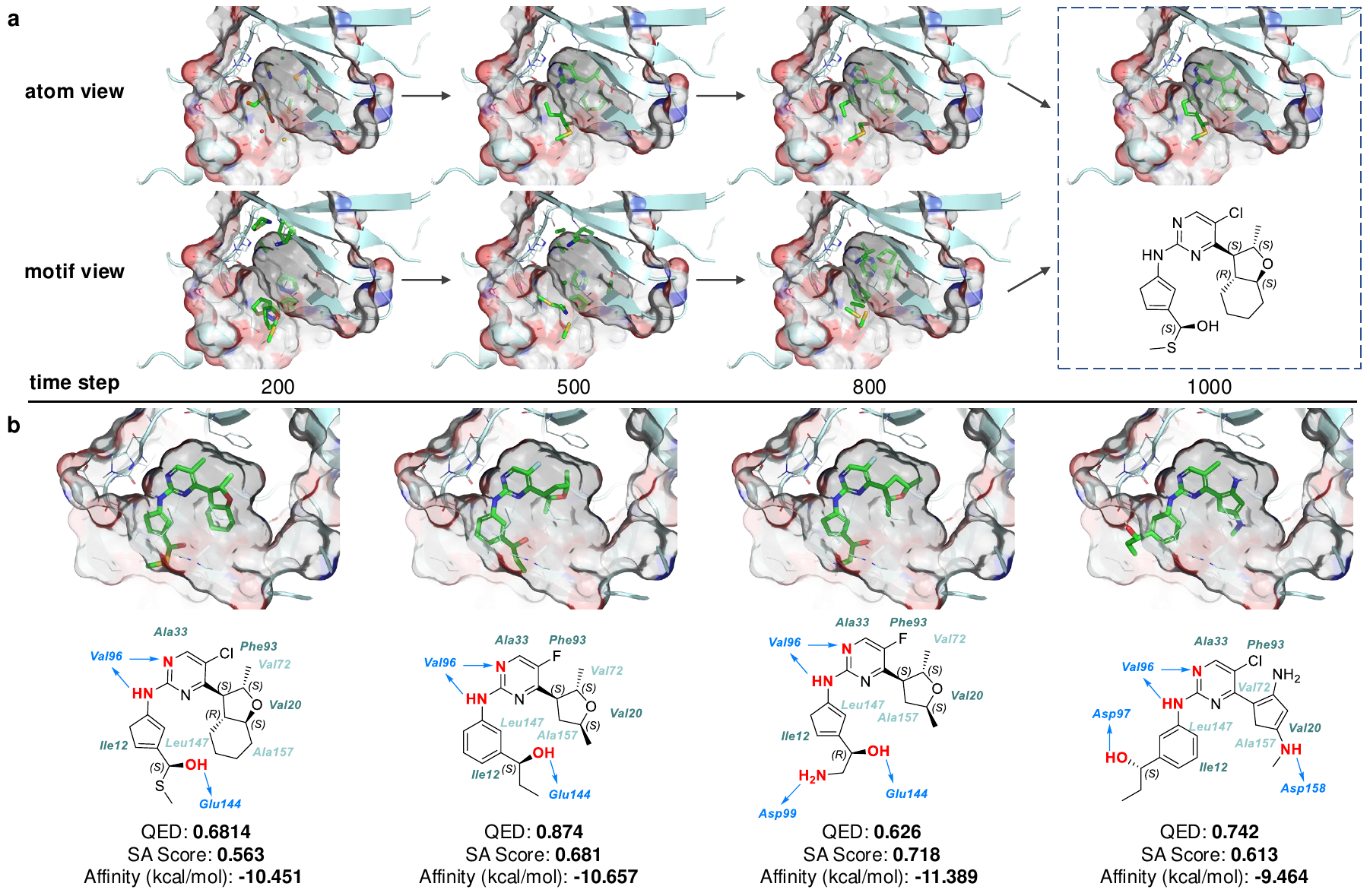}
\caption{ 
Examples of Molecules Generated by AMDiff targeting CDK 4 (PDB id:7SJ3). \textbf{(a)} An example of a conditional design trajectory . At initial time steps, substructures progressively explore interactions with the pocket in both atom-view and motif-view. The trajectory gradually refines into a realistic molecule structure. \textbf{(b)} Molecules designed to target CDK 4 (PDB id:7SJ3), with molecular properties such as QED and SA score, as well as binding affinity and protein-ligand interaction analysis.}\label{result4}
\end{figure*}
Among the affinity prediction (docking score), AMDiff demonstrates the highest performance, indicating that our model has effectively learned the favorable molecular conformations within the target pockets. Additionally, we assess the molecular properties of the generated compounds. AMDiff achieves the highest QED and SA values, which closely resemble the distribution of active compounds. Figure \ref{result2} (e) displays the heatmaps of the docking score and QED value distributions for molecules generated by Pocket2Mol, FLAG, and AMDiff. Each data grid is color-coded according to the corresponding SA score. Molecules generated using AMDiff exhibited higher docking scores compared to those generated using Pocket2Mol and FLAG, and were in proximity to the docking scores of bioactive ligands.
In addition to the aforementioned indicators, we also assess the spatial similarity between the molecular poses directly generated by models and those obtained after molecular docking. Specifically, we investigate the root-mean-square deviation (RMSD) between the generated conformations and docked structures, with detailed definitions provided in Equation \ref{eq:rmsd} of the Methods section \ref{model_evalutation}. As shown in Figure \ref{result2} (f), our model exhibits lower RMSD values, indicating minimal deviation from the optimal docked conformations. This demonstrates AMDiff's ability to generate conformations with minimal shifts, closely aligning with the docked poses.

To verify the capability of AMDiff in recognizing protein pockets, we further validated it using 3D visualization. Firstly, to visualize the generative process of the AMDiff, we showcase the gradual generative diffusion process of the model within the binding pocket of CDK4, as in Figure \ref{result4} (a). At nodes with time steps of 200, 500, 800, and 1000, atom-view and motif-view progressively capture the features of the protein pockets and guide the diffusion process. Through generation and interaction at these nodes, compound 1 is ultimately formed. AMC-Diff demonstrates distinct generative processes in both atom- and motif-views, allowing for observations of atom or motif substitutions. However, following the cross-view interaction within the hierarchical diagram, the final molecule integrates the benefits of both views, resulting in a structure that is well-suited to the protein pocket. 
Then, based on similar molecular generation processes, we screen out other compounds from the generated molecules as potential CDK4 inhibitors, and analyze whether our model effectively learns the intricate microscopic interaction patterns within protein-ligand complexes. Key molecular descriptors are exhibited, including the quantitative estimate of QED, SA scores and top-1 docking score from AutoDcok Vina. The best conformation for each compound is also described in both 2D and 3D view. As shown in Figure \ref{sec4} (b), most of these molecules exhibit interactions with the same amino acid residues. This suggests that the generated molecules are capable of fitting into the binding sites. Regarding pharamacophoric groups, AMDiff creates common important pharmacophore elements as the reference ligands. Specifically, compounds 1 and 2 form hydrogen bonds with Val96 and Glu144. Compound 3 forms hydrogen bonds with Val96, Glu144, and Asp99, as well as a pi-cation interaction with Asp99. Compound 4 forms hydrogen bonds with Val96, Asp97, and Asp158. The binding modes of the generated compounds align with the recognized binding patterns, demonstrating the target-aware ability of AMDiff to utilize known interactions while potentially uncovering novel ones.

\vspace{-3mm}

\subsection{AMDiff exhibits robustness in protein evolution, adapting to mutated proteins and multi-scale pocket sizes}

\vspace{-1mm}
 
Protein mutations are a natural part of evolution but can often lead to drug resistance during treatment, as they alter target interactions and modify pocket shape and size. Given AMDiff's performance in the gradual diffusion process, we explored its 

\begin{figure}[H]%
\centering
\includegraphics[width=1\textwidth]{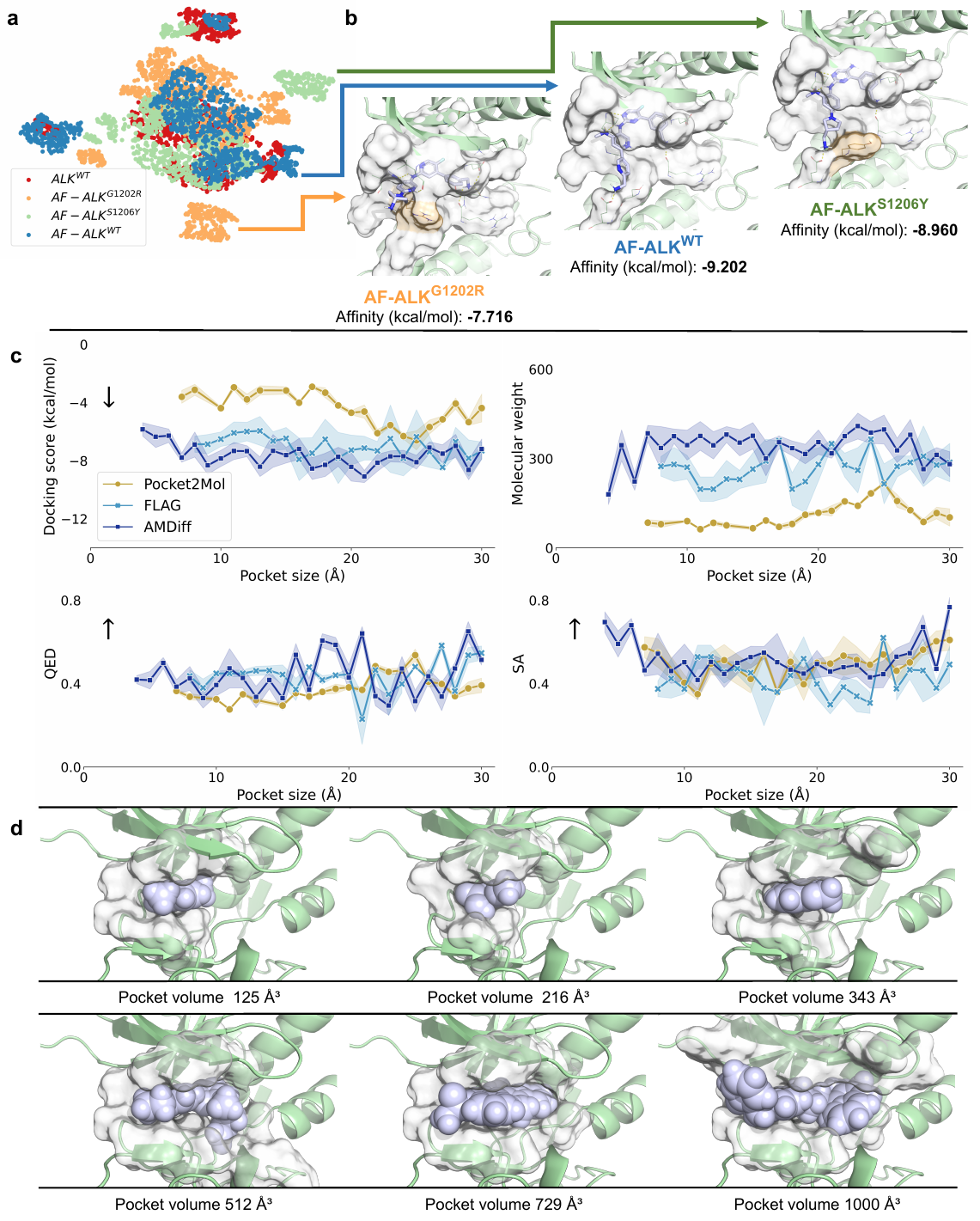}
\caption{\textbf{(a)} The distribution of molecules generated after mutating ALK (PDBID: 3LCS) is shown. The clustering results of USRCAT fingerprints for molecules targeting three mutations were visualized using t-SNE in two-dimensional space. $\operatorname{ALK^{WT}}$: Wild-type ALK proteins form PDB bank (PDB id: 3LCS). $\operatorname{AF-ALK^{WT}}$: Wild-type ALK proteins form Alphafold (PDB id: 3LCS). $\operatorname{AF-ALK^{G1202R}}$: A substitution of the amino acid Gly with Arg at position 1202 in the protein sequence. $\operatorname{AF-ALK^{S1206Y}}$: A substitution of the amino acid Ser with Tyr at position 1206 in the protein sequence. \textbf{(b)} Examples of molecules generated after modifying residues within the pocket of $\operatorname{AF-ALK^{G1202R}}$, $\operatorname{AF-ALK^{S1206Y}}$ and $\operatorname{AF-ALK^{WT}}$. The mutated  \textbf{(c)} Conditional generation of molecules for various pocket sizes targeting ALK (PDB ID: 3LCS). Comparison of key property performance when utilizing binding pockets of varying sizes, including docking score, molecular weight, SA and QED. \textbf{(d)} Visualization examples showcasing generated samples adjusted to match different pocket sizes. The molecular volumes are tailored to correspond with the given pocket volumes.}\label{result5}
\end{figure}

\noindent capability to address mutant proteins. The ALK protein mutation is a well-known challenge in drug development, as such mutations can significantly reduce therapeutic efficacy.
To address this, we generate ALK inhibitors against both wild-type ALK proteins from PDB bank ($\operatorname{ALK^{WT}}$, PDB id: 3LCS) and Alphafold \cite{jumper2021highly} ($\operatorname{AF-ALK^{WT}}$). We also design two mutant proteins based on $\operatorname{AF-ALK^{WT}}$ through site-directed mutagenesis: (i) $\operatorname{AF-ALK^{G1202R}}$, where glycine (Gly) at position 1202 is replaced with arginine (Arg), and (ii) $\operatorname{AF-ALK^{S1206Y}}$, where serine (Ser) at position 1206 is substituted with tyrosine (Tyr). Figure \ref{result4} (c) present the t-SNE visualization of the distributions of the USRCAT fingerprint \cite{schreyer2012usrcat} for molecules generated for these four proteins. The results reveal significant overlap in the chemical space of the generated molecules, yet there are distinct regions where the distributions do not overlap. This indicates that AMDiff can explore variations in local areas and align generated molecules with the target binding sites effectively. We further showcase the 3D-binding modes and molecular differences by AMDiff in Figure \ref{result4} (d). These molecules can form robust hydrogen bonds with Met1199 on $\operatorname{AF-ALK^{WT}}$, $\operatorname{AF-ALK^{G1202R}}$ and $\operatorname{AF-ALK^{S1206Y}}$. For mutations at positions 1202 and 1206, AMDiff recognizes steric hindrance and generates differentiated molecular structures in a targeted manner, coordinating energy loss at the global level. Given AMDiff's sensitivity to mutagenesis-induced differences during molecular generation, we further investigate its performance at different scales of protein pockets. Specifically, we assess ligands generated within ALK pocket sizes ranging from 4 Å to 30 Å.  Figure \ref{result5} (a) compares the docking score, molecular weight, QED, and SA score of these molecules with those generated by the Pocket2Mol and FLAG. Figure \ref{result5} (b) performs the 3D molecular spheres illustrating pocket fitness for molecules generated by AMDiff. The results indicate that AMDiff successfully generated viable molecules across all pocket scales. In contrast, Pocket2Mol and FLAG exhibite limitations in generating normal molecules when the binding sites were too small (4 Å), suggesting these methods are constrained by their preset pocket boundaries. This could be attributed to the introduction of atom-view in AMDiff, which can construct smaller pieces when encountering larger hinderances, thus accommodating small pocket size.

\section{Discussion}\label{sec4}
\subsection{Hierarchical Representation}

Hierarchical organization is a fundamental characteristic of numerous biological structures, including proteins, DNA, and molecules. Hierarchical representations have been successfully employed in various models, offering a more comprehensive understanding of biological systems at multiple scales. Protein structures can be represented at various levels of granularity, including
the amino acid, backbone, and all-atom levels. ProNet \cite{wang2022learning} capture hierarchical relations among different levels and learn protein representations. HierDiff \cite{qiang2023coarse} employs a diffusion process to generate fragment representations instead of deterministic fragments in the initial step, followed by constructing atom-level 3D structures. HIGH-PPI \cite{gao2023hierarchical} establish a robust understanding of Protein-Protein Interactions (PPIs) by creating a hierarchical graph that encompasses both the PPI graph and protein graph. By incorporating multiple levels of structural information, these models offer more nuanced and comprehensive representations of biological systems, potentially leading to improved predictions and deeper insights into biological functions and interactions.

In this paper, we tackle the de novo ligand design problem from a hierarchical perspective, introducing a cross-view diffusion model that generates molecular structures at both the atomic and motif views. Our model excels in recognizing multi-level geometric and chemical interactions between ligands and target proteins. By capturing varying levels of granularity, we construct ligand-protein and cross-view interactions. Existing methods often neglect or inadequately utilize the hierarchical structure within molecules. Through empirical evaluation, our model, AMDiff, demonstrates its strong capability to generate valid molecules while maintaining the integrity of local segments. Furthermore, it exhibits robustness across diverse protein structures and pocket sizes.

\subsection{Limitations and future work}

We present three primary limitations of AMDiff and propose potential solutions for future research. Firstly, we do not explore molecule generation in dynamic systems. In real-world applications, protein structures can undergo conformational changes, leading to shape shifts in pockets. Additionally, the formation of cryptic pockets expands the possibilities for drug discovery by enabling the targeting of proteins with multiple druggable binding sites. We recommend that future work considers the dynamics of protein structures, accounting for intrinsic or induced conformational changes. Secondly, it is essential to incorporate more domain knowledge into the model-building process. The integration of chemical and biomedical prior knowledge has proven effective in various tasks. Investigating the interactions between proteins and bioactive ligands, such as hydrogen bonds, salt bridges, pi-cation, and pi-pi stacking, is crucial. Furthermore, exploring the influence of pharmacophore elements during the generation of bioactive molecules would be valuable. Lastly, while we evaluate the designed drug candidates using multiple metrics, it remains necessary to tightly collaborate with medicinal chemists and conduct wet-lab experiments for in vitro or in vivo validation of their effectiveness. The experimental results obtained can then be utilized to refine and improve the generative model.

\section{Methods}\label{sec4}

The protein can be represented as a set of atoms $P = \{ (\bm x^{(i)}, \bm v^{(i)}) \}_{i=1}^{N_P}$ where $N_P $ is the number of protein atoms, $\bm x \in \mathbb R^{3}$ represents the 3D coordinates of the atom, and $\bm v \in \mathbb R^{F}$ represents protein atom features such as amino acid types. The molecule with N atoms can be represented as $G = \{ (\bm x^{(i)}, \bm v^{(i)}) \}_{i=1}^{N_G}$ where $\bm x \in \mathbb{R}^{3}$ is the atom coordinate, $\bm v \in \mathbb{R}^{V}$ is the one-hot atom type. $R = \{ (\bm x^{(i)}, \bm v^{(i)}) \}_{i=1}^{N_R}$ is the atom 3D coordinates and features of binding pocket. 
The goal is to develop a generative model, denoted as $p(G|R,P)$, that predicting the three-dimensional structure of ligands conditioned on the protein and binding pocket.

\subsection{Classifier-free guidance diffusion model}\label{model_diff}


We develop a conditional diffusion model for target-specific molecule generation. Our approach consists of a forward diffusion process and a reverse generative process. The diffusion process gradually injects noise to data, and the generative process learns to reconstruct data distribution from the noise distribution using a network parameterized by $\theta$. The latent variable of every time step is represented as $p_\theta(G_t|(R,P))$, indicating that the predicted molecules are conditioned on the provided protein and binding sets. We can represent the diffusion process and the recovery process as follows:

\begin{equation}
\begin{aligned}
    q  \left( G_{1:T}|G_0,(R,P) \right) & = \prod_{t=1}^{T} q \left( G_t|G_{t-1},(R,P) \right), \\
    p_\theta \left( G_{0:T-1}|G_T,(R,P) \right) &= \prod_{t=1}^T p_\theta \left( G_{t-1}|G_t,(R,P) \right),
\end{aligned}
\end{equation}

\noindent where $G_1, G_2, \cdots , G_T$ is a sequence of latent variables of with the same dimensionality as the input data $G_0 \sim p(G_0|R,P)$. 
Following continuous diffusion model \cite{ho2020denoising}, during the forward process, we add Gaussian noise to atom coordinates at each time step $t$ according to the variance schedule $\beta_1,...,\beta_T$:
\begin{equation}\label{eq:diff}
    q(\bm x_t|\bm x_0) = \mathcal{N}(\bm x_t; \sqrt{\bar \alpha_t}\bm x_0, (1-\alpha_t) \operatorname{\bm I}),  \quad \alpha_t := 1-\beta_t \quad \bar{\alpha}_t:=\Pi_{s=1}^t \alpha_s.
\end{equation}
The model can be trained by optimizing the KL-divergence  between $q(\bm {x}_{t-1}|\bm x_t, \bm x_0)$ and $p_\theta (\bm  x_{t-1}|\bm x_t)$.
The posterior distributions of the forward process, $q(\bm x_{t-1}|\bm x_t, \bm x_0)$ can be represented as:
\begin{equation}
\begin{aligned}
& q(\bm x_{t-1}|\bm x_t, \bm x_0) = \mathcal{N}(\bm x_{t-1}; \tilde{\mu}_t(\bm x_t, \bm x_0), \tilde{\beta}_t \operatorname{\bm I}), \\
& \tilde{\mu}_t(\bm x_t, \bm x_0):= \frac{\sqrt{\bar{\alpha}_{t-1}}\beta_t}{1-\bar{\alpha}_t}\bm x_0 + \frac{\sqrt{\alpha_t}(1-\bar{\alpha}_{t-1})}{1-\bar{\alpha}_t} \bm x_t, \quad \tilde{\beta}_t := \frac{1-\bar{\alpha}_{t-1}}{1-\bar{\alpha}_t} \beta_t.
\end{aligned}
\end{equation}
In the denoising process, the transition distribution can be written as $p_\theta(\bm x_{t-1}|\bm x_t) = \mathcal{N}(\bm x_t-1; \mu_\theta(\bm x_t, t), \sigma^2_t \operatorname{\bm I})$ and let $\sigma_t^2=\beta_t$ experimentally. To reparameterize Equation \ref{eq:diff}, we define $\bm x_t(\bm x_0, \bm \epsilon) = \sqrt{\bar{\alpha}_t} \bm x_0 + \sqrt{1-\bar{\alpha}_t } \bm \epsilon$ where $\bm \epsilon \sim \mathcal{N} (0,\operatorname{\bm I})$. Consequently, the final objective of the training process can be expressed as:
\begin{equation}
L_{\operatorname{pos}} = \mathbb{E}_{\bm x_0, \bm \epsilon} \left [\frac{\beta_t^2}{2\sigma_t^2 \alpha_t(1-\bar{\alpha}_t)} \parallel \bm \epsilon - \bm \epsilon_\theta(\sqrt{\bar{\alpha}_t }\bm x_0 + \sqrt{1-\bar{\alpha}_t} \bm \epsilon,t,(R,P))\parallel^2 \right ].
\end{equation}

The representation of atom types can be achieved through a one-hot vector, denoted as $\bm v_t$. To predict the atom types in molecules, we utilize a discrete diffusion model \cite{hoogeboom2021argmax,austin2021structured}.  At each timestep, a uniform noise term, $\beta_t$, is added to the previous timestep $\bm v_{t-1}$ across the $K$ classes. This discrete forward transition can be expressed as follows:
\begin{equation}
    q(\bm v_t|\bm v_{t-1}) = \mathcal{C}(\bm v_t; (1- \beta_t) \bm v_{t-1}+\beta_t/K),
\end{equation}
where $\mathcal{C}$ is the categorical distribution. Through Markov chain, we can get:
\begin{equation}
    q(\bm v_t|\bm v_0) = \mathcal{C} (\bm v_t | \bar{\alpha}_t \bm v_0 + (1-\bar{\alpha}_t)/K),
\end{equation}
where $\alpha_t = 1 - \beta_t$ and $\bar{\alpha}_t = \prod_{\tau = 1}^t \alpha_\tau$. Then the categorical posterior can be calculated as:
\begin{equation}
\begin{aligned}
 & q(\bm v_{t-1}|\bm v_t, \bm v_0) = \mathcal{C} (\bm v_{t-1}| \bm Q(\bm v_t, \bm v_0)), \quad \bm Q(\bm v_t, \bm v_0) = \tilde{G}/\sum_{k=1}^K \tilde{G}_k, \\
 & \tilde{G} = [\alpha_t \bm v_t + (1-\alpha_t)/K] \odot [\bar{\alpha}_{t-1} \bm v_0 + (1-\bar{\alpha}_{t-1})/K].
\end{aligned}
\end{equation}

For the training object of atom types, we can compute KL-divergence of categorical distributions:

\begin{equation}
L_{type}  = \sum \limits_k \bm Q( \bm v_t, \bm v_0)_k \operatorname{log} \frac{\bm Q(\bm v_t,\bm v_0)_k}{\bm Q(\bm v_t,\hat{\bm v}_0)_k}.
\end{equation}

Motivated by the ability of guided diffusion models to generate high quality conditional samples, we apply classifier-free guided diffusion to the problem of pocket-conditional molecule generation. The feature of pocket provides a useful guidance signal to generation 3D structure in binding sites. During training process, the pocket in the diffusion model $\bm \epsilon_\theta (G_t|R,P)$ is replaced with a null label $\emptyset$ with a fixed probability. During sampling, the output of the model is extrapolated further in the direction of $\bm \epsilon_\theta (G_t|R,P)$ and away from $\bm \epsilon_\theta (G_t|\emptyset,P)$:

\begin{equation}
\title{ \hat{\bm \epsilon}}_\theta (G_t\mid (R,P)) = \bm \epsilon_\theta (G_t \mid (\emptyset ,P)) + s \cdot \big ( \bm \epsilon_\theta (G_t (R,P)) - \bm \epsilon_\theta (G_t \mid (\emptyset ,P))  \big ).
\end{equation}

\noindent Here, $s \geq 1$ is the guidance scale. By jointly train a conditional and an unconditional diffusion model, we combine the resulting conditional and unconditional score estimates to attain a trade-off between sample quality and diversity similar to that obtained molecules using classifier guidance. This methodology proves useful in obtaining a truncation-like effect across various proteins and sets of multi-druggable bindings.

\subsection{Joint training for hierarchical diffusion}\label{model_Hierarchical}

We build the Atom-Motif Consistency Diffusion Model (AMDiff), which is implemented in a joint training manner. This model incorporates both atom view and motif view to construct molecular structures. The atom view utilizes atoms as fundamental building blocks, allowing for the generation of highly diverse molecular structures. On the other hand, the motif view assembles subgraphs by leveraging fragments within a motif vocabulary, aiding in the learning of prior patterns. The interaction between these two views promotes transfer learning.

A motif $M_i$ is defined as a subgraph of a molecule $G$.
Given a molecule, we extract its motifs $M = \{ (\bm x^{(i)}, \bm w^{(i)}) \}_{i=1}^{N_M}$, where $\bm x \in \mathbb{R}^{3}$ represents the motif coordinates in 3D space, $\bm w \in \mathbb{R}^{W}$ denotes the motif IDs in the motif vocabulary. We employ atom-view and motif-view diffusion models to generate feature representations in a latent space.
To formulate the proposed AMDiff approach, we introduce a hierarchical diffusion model for the atom view and motif view, denoted by $\Phi(G_t,\theta_1)$ and $\Phi(M_t,\theta_1)$, respectively. These hierarchical diffusion networks update simultaneously, and the overall recovery process can be represented as: 

\begin{equation}
(\hat{G}_0, \hat{M}_0)=\Phi_{\theta_1,\theta_2}(G_t,M_t,t, (R,P)).
\end{equation}

The reverse network, denoted as $\Phi_{\theta_1,\theta_2}(G_t,M_t,t, (R,P))$, is constructed using equivariant graph neural networks (EGNNs) \cite{satorras2021n}. We build the k-nearest neighbors graph between ligand atoms, motifs with the condition pocket and protein atoms, and using message passing to model the interaction between them:
\begin{equation}
\begin{aligned}
    \operatorname{Mes}_{i,j}  = \phi_{ \operatorname{Mes}} (\bm h^l_i, \bm h^l_j, \bm e^l_{ij}, t, \parallel \bm x_i -\bm x_j \parallel), \quad &
    \bm h^{(l+1)}_i  = \phi_{f} (\bm h^l_i, \sum_{j \in \mathcal{N}(i)}  \operatorname{Mes}_{ij}). 
    \\
   \Delta \bm x^{l+1}_i = \phi_x (\bm h^{(l+1)}_i), \quad &  \bm x^{l+1}_i = \bm x^l_i + \Delta \bm x^{l+1}_i. 
\end{aligned}
\end{equation}

\noindent where $\bm h_i$ and $\bm h_j$ represent for the embeddings of vertices in the constructed neighborhood graph. The variable $\bm e_{ij}$ indicates the edge type, which is determined by the type of vertices connected by the edge. The variable $l \in (0,L)$ represents the index of the EGNN layer, where $L$ denotes the total number of layers. $\bm x_i$ and $\bm x_j$ represent the coordinates of the nodes. The networks $\phi_{Mes}$, $\phi_{f}$, and $\phi_{x}$ are utilized for computing the message, feature embedding, and position, respectively. After get the position of every node, we calculate the atom type and motif id using the type prediction networks $\phi_v$ and $\phi_w$:
\begin{equation}
 \bm v_G = \phi_v (\bm h^L_G), \quad \bm w = \phi_w (\bm h^L_M).
\end{equation}

\subsection{Topological features}\label{model_topological}

The  geometric  and  topological  information  existed  in  molecule  structure  is  the  essential  clue  to  understand the interaction between drug and protein. However, traditional general GNNs models often struggle to capture the intricate structures of 3D objects, such as cycles and cavities. To address this limitation, we employ persistent homology, a mathematical tool utilized in topological data analysis \cite{meng2021persistent,liu2023persistent,nguyen2020review}, to extract stable topological features from pocket and ligand point clouds. 

We utilize filtration functions, denoted as $f$, to calculate the persistence diagrams (PD) $\operatorname{ph}(\bm x,f) = \{\bm {D}_0, ..., \bm {D}_l\}$, where $\bm {D}_l$ is the $l$-dimensional diagram and x is the point cloud of pockets or ligands. The resulting PD reflects the multi-scale summarized topological information.
Then we calculate the normalized persistent entropy from the PD through the method introduced in \cite{myers2019persistent}: 
\begin{equation}
\begin{aligned}
    E(\bm D_l)  = -\sum_{d \in \bm D_l} \frac{\operatorname{pers(d)}}{\mathcal{E}(\bm D_l)} \operatorname{log}_2 & \left (\frac{\operatorname{pers(d)}}{\mathcal{E}(\bm D_l)} \right ),  \quad   \mathcal{E} (\bm D_l) = \sum_{d \in \bm D_l} \operatorname{pers}(d), \\
    E_{\operatorname{norm}} (\bm D_l) = & \frac{E(\bm D_l)}  {\operatorname{log}_2(\mathcal{E}(\bm D_l))}.
\end{aligned}
\end{equation}
where $d$ represents the persistence of points in the diagram $\bm D_l$, and $\operatorname{pers}(d)$ denotes the lifetime. The normalized persistent entropy $E_{\operatorname{norm}}(\bm D_l)$ quantifies the entropy of the persistence diagram.
We use an embedding function $\phi_d$ to map persistence diagrams into vector representations in high-dimensional space $\phi_d:\{ \bm D_1,...,\bm D_l \} \rightarrow \mathbb R ^{n' \times d}$, where $n'$ is the dimension of vector. Consequently, we can get the topological fingerprints of ligands and pockets:
\begin{equation}
\bm F_G = \phi_d(E_{\operatorname{norm}} \left ( \operatorname{ph}({\bm x_G})\right )), \quad \bm F_R  = \phi_d(E_{\operatorname{norm}} \left ( \operatorname{ph}({\bm x_R})\right )).
\end{equation}

The topological fingerprints is incorporated to pocket and ligand representations.
Topological constraints are appended in optimization to enhance the coherency and connectivity in molecule output.

\subsection{Training Strategy}

We construct a hierarchical structure consisting of atom-view and motif-view representations during the diffusion process. To train our model, we adopt a joint training approach that provides supervision for each view and their parameters updated simultaneously by a joint loss. In the atom-view, we employ two types of losses as introduced in \ref{model_diff}: the position loss ($L_{\operatorname{a_{pos}}}$) and the atom type loss ($L_{\operatorname{a_{type}}}$).
Similarly, in the motif-view, we incorporate the motif position loss ($L_{\operatorname{m_{pos}}}$) and the motif id loss ($L_{\operatorname{m_{id}}}$). Consequently, the final loss is determined as a weighted sum of the coordinate loss and type loss, expressed as follows:

\begin{equation}
L = L_{\operatorname{a_{pos}}} + \lambda_1 L_{\operatorname{a_{type}}} +  L_{\operatorname{m_{pos}}} + \lambda_2 L_{\operatorname{m_{id}}}.
\end{equation}

\noindent Here, $\lambda_1$ and $\lambda_2$ are parameters used to control the contribution of the loss terms in the overall loss function.

\subsection{Medicinal Chemistry Filters}\label{model_MCF}

We utilize filters to refine our search process for preliminary candidates with favorable medicinal chemistry and structural novelty. 
First, we utilize structural filters to eliminate structures with problematic substructures. These substructures encompass promiscuously, reactive substructures (like pan-assay interference compounds, PAINS), pharmacokinetically unfavorable substructures (Brenk substructures) \cite{baell2010new,brenk2008lessons}, and other alerts.
We employ these filters to screen compounds and exclude those with potentially toxic structures and compounds containing undesirable groups.  
Second, we employ property filters to further refine our selection. These filters aim to eliminate compounds that are unlikely to exhibit optimal molecular properties, thus enhancing the quality of our candidate pool. 
Finally, we leverage the Tanimoto similarity metric to assess the resemblance of our compounds to bioactive molecules. Specifically, we calculate Tanimoto similarity scores to identify compounds with a high degree of similarity to known ALK ligands. By prioritizing compounds that demonstrate significant Tanimoto similarity to these reference ligands, we increase the likelihood of identifying candidates with potential targeted activity against ALK. 

\subsection{Datasets}

Following the works of \cite{luo20213d} and \cite{zhang2022molecule}, we utilize the CrossDocked dataset \cite{francoeur2020three} to train and evaluate our model. This dataset comprises a comprehensive collection of 22.5 million docked protein binding complexes. We drop all complexes that cannot be processed by the RDKit and filter out complexes with a binding pose Root Mean Square Deviation (RMSD) greater than 2 \r{A} and a sequence identity lower than 40\%.  This filtering step aimed to remove complexes displaying significant structural deviations and low sequence similarity. After applying these filters, we obtain a subset of 100,000 complexes for training our model, while reserving an additional 100 proteins for testing purposes. For real-word therapeutic targets, we download the protein structures form PDB \cite{sussman1998protein}, and the active molecules corresponding to these targets were downloaded from BindingDB \cite{liu2007bindingdb}.  

\subsection{Evaluation}\label{model_evalutation}

In this work, we apply the widely used metrics of deep generative models to evaluate the performance of our method. (1) \textbf{Validity} measures the percentage of generated molecules that successfully pass the RDKit sanitization check.
(2) \textbf{Uniqueness} is assessed by calculating the percentage of unique structures among the generated outputs. 
(3) \textbf{Diversity} considers the proportion of unique scaffold structures generated, as well as the internal diversity values calculated for 2D and 3D structures using the Morgan fingerprint and USRCAT, respectively.
(5) \textbf{Novelty} measures the proportion of unique scaffold structures generated that are not present in either the test set or the known ligand set.
(6) \textbf{Molecular Properties}. We report the distributions of several important 2D and 3D molecular properties, including molecular weight (MW), QED, Synthesizability (SA), log P, normalized principal moment of inertia ratios (NPR1 and NPR2). These distributions are compared to those of the train and test set to assess the model's ability to learn important molecular properties. (6) \textbf{Affinity}: The average binding affinity of the generated molecules, which is assumed to be characterized by the docking score. The value in kcal/mol is estimated by AutoDock Vina. (7) \textbf{RMSD} stands for root-mean-square deviation, which is a metric used to measure the dissimilarity between different conformations of the same molecule. A smaller RMSD value indicates a higher degree of similarity between the conformations. The formula for calculating RMSD is as follows:

\begin{equation}\label{eq:rmsd}
\operatorname{RMSD}(\hat{\bm R}, \bm R) = \left(\frac{1}{n} \sum_{i=1}^{n} \parallel \hat{\bm R}_i - \bm R_i  \parallel^2 \right)^{\frac{1}{2}}.
\end{equation}

\noindent Here, $n$ represents the number of atoms, $\hat{\bm R}$ represents the conformation of the generated molecule, and $\hat{\bm R_i}$ represents the Cartesian coordinates of the i-th atom. We use the RMSD calculation to assess the conformational changes that occur before and after the docking process.

We conducte a comparative analysis with several baselines: 
\textbf{liGAN} \cite{ragoza2022generating} is a method that combines a 3D CNN architecture with a conditional variational autoencoder (VAE), enabling the generation of molecular structures using atomic density grids.  
\textbf{AR} \cite{luo20213d} introduces an auto-regressive sampling scheme to estimates the probability density of atom occurrences in 3D space. The atoms are sampled sequentially from the learned distribution until there is no room within the binding pocket. 
\textbf{Pocket2Mol} \cite{peng2022pocket2mol} design a new graph neural network capturing both spatial and bonding relationships between atoms of the binding pockets, then samples new drug candidates conditioned on the pocket representations from a tractable distribution. 
\textbf{GraphBP} \cite{liu2022generating} use a flow model to generate the type and relative location of new atoms. It also obtains geometry-aware and chemically informative representations from the intermediate contextual information. 
\textbf{DecompDiff} \cite{guan2024decompdiff} is an end-to-end diffusion-based method that utilizes decomposed priors and validity guidance to generate atoms and bonds of 3D ligand molecules.
\textbf{FLAG} \cite{zhang2022molecule} constructs a motif vocabulary by extracting common molecular fragments from the dataset. The model selects the focal motif, predicts the next motif type, and attaches the new motif.

\section{Conclusion}\label{sec5}

In this work, we present AMDiff, a novel deep learning approach for \textit{de novo} drug design. Our method focuses on generating 3D molecular structures within the binding site of a target protein. By employing the conditional diffusion model, AMDiff effectively learns the interactions across various binding pockets of the protein. 
One distinct feature of AMDiff is its hierarchical structure within the diffusion process. This architecture allows us to capture the 3D structure at both the atom and motif levels, overcoming the limitations of conventional atom- and motif-based molecular generation approaches. We incorporate a classifier-free guidance mechanism that provides interaction signals randomly during each denoising step, enabling the network to strengthen its guidance in identifying pocket structures. 

To evaluate the performance of AMDiff, we conduct experiments aimed at designing potential hits for selected drug targets, using ALK and CDK4 as case studies. The results demonstrate that AMDiff can successfully generates drug-like molecules with novel chemical structures and favorable properties, exhibiting significant pharmacophore interactions with the target kinases. Additionally, AMDiff demonstrates high flexibility in generating 3D structures with various user-defined pocket sizes, further enhancing its utility. Overall, our work introduces a promising tool for structure-based de novo drug design, with the potential to significantly accelerate the drug discovery process.

\bmhead{Data availability} 

The train and test dataset CorssDocked is available at 

\noindent \href{https://drive.google.com/file/d/1kjp3uLft4t6M62HgSAakiT7BnkQaSRvf/view?usp=sharing}{https://drive.google.com/file/d/1kjp3uLft4t6M62HgSAakiT7BnkQaSRvf/view?usp=sharing}. Instructions and data loaders are provided via the associated GitHub repository. Case studies targeting protein kinases, including Anaplastic Lymphoma Kinase (ALK) and Cyclin-dependent kinase 4 (CDK4), can be found on the RCSB Protein Data Bank (PDB) website at \href{https://www.rcsb.org/}{https://www.rcsb.org/}

\bmhead{Code availability}
The source code for model architecture is publicly available on  our GitHub repository (\href{https://github.com/guanlueli/AMDiff}{https://github.com/guanlueli/AMDiff}).

\backmatter




\bigskip












\bibliography{sn-bibliography}

\end{document}